\documentclass[10pt,twocolumn,letterpaper]{article}

\usepackage{adjustbox}
\usepackage{commath}
\usepackage{cvpr}
\usepackage{times}
\usepackage{epsfig}
\usepackage{graphicx}
\usepackage{amsmath}
\usepackage{amssymb}
\usepackage{enumitem}
\usepackage{xcolor}
\usepackage[colorlinks = true,
            linkcolor = blue,
            urlcolor  = blue,
            citecolor = blue,
            anchorcolor = blue]{hyperref}
            
\cvprfinalcopy 

\def\cvprPaperID{****} 

\begin{document}

\title{Image-based Virtual Fitting Room}

\author{Jie Chen\thanks{equal contribution}, Junwen Bu*, Zhiling Huang*\\
Department of Computer Science, Stanford University\\
{\tt\small \{jiechen8, junwenbu, zhiling\}@stanford.edu}
}

\maketitle

\begin{abstract}
   Virtual fitting room is a challenging task yet useful feature for e-commerce platforms and fashion designers. Existing works can only detect very few types of fashion items. Besides they did poorly in changing the texture and style of the selected fashion items. In this project, we propose a novel approach to address this problem. We firstly used Mask R-CNN \cite{maskrcnn} to find the regions of different fashion items, and secondly used Neural Style Transfer \cite{DBLP:journals/corr/GatysEB15a} to change the style of the selected fashion items. The dataset we used is composed of images from PaperDoll \cite{paperdollgithub} dataset and annotations provided by eBay's ModaNet \cite{ModaNet}. We trained 8 models and our best model massively outperformed baseline models both quantitatively and qualitatively, with 68.72\% mAP, 0.2\% ASDR.
\end{abstract}
\section{Introduction}
Imagine you try new clothes by using Virtual Fitting Room on your cell phone. The app can deliver a virtual try on experience as if you are in a fitting room of Macy's. You upload a photo of portrait. Then you select the fashion items you want to change and also new styles for those selected fashion items. The App will generate a new image, with the selected fashion items substituted with new ones that have the style you want.

The Virtual Fitting Room is a challenging task yet useful feature for e-commerce platforms and fashion designers. There are three key issues that make the problem difficult. Firstly there can be a big number of fashion items that are of different categories in one single fashion model image. Secondly it is hard to find a perfect segmentation for each fashion item. Thirdly it is even more difficult to change the fashion item into an arbitrary style.

Our project focuses on swapping selected fashion items on a portrait image with fashion items that are generated by deep neural network. \textbf{The input to our algorithm consists of two images: 1) a portrait image with fashion items; 2) a texture image. We then use a Mask R-CNN network \cite{maskrcnn} and a Neural Style Transfer network \cite{DBLP:journals/corr/GatysEB15a} to output a new portrait image where the style of the selected fashion items are changed to the input texture.}

For example, given a portrait image, pants texture can be changed from canvas to jeans; coat texture can be changed from cotton to leather. All these are depended on the input texture images. We can adopt different combinations depending on user's preference in different scenarios: day or night; indoor or outdoor with different body poses.
\section{Related Work}
There is a large body of work trying to achieve goals similar to ours. After thoroughly reviewing previous works, we grouped them into three sub-categories: 1) Recognition and segmentation for fashion items, 2) Clothing Styles \& Texture Switching, 3) Virtual Try-on.
\subsection{Recognition \& Segmentation of Fashion Items}
 Extensive studies have been conducted on semantic image recognition and segmentation in the context of fashion due to its huge profit potentials. \par
 For clothing recognition, early work such as A. Borras et al. \cite{borras} attempted to identify layers of upper body clothes in very limited situation. More recently work L. Bourdev et al. \cite{6126413} attempted to consider clothing items as semantic attributes of a person, but only limited to a small number of garments. Different from these approaches, K. Yamaguchi \cite{Yamaguchi:Clothing} estimated a complete and precise region based labeling with a relatively large number of potential garment types and further improved the performance by using retrieval-based approach \cite{article:Retrieving}.\par
 For clothing segmentation, Shotton et al. \cite{inproceedings:Shotton} proposed an approach mainly based on region-based segmentation and object detection. K. Yamaguchi \cite{Yamaguchi:Clothing} took a related approach (CRF based labeling) and focused on estimating labeling for several types of fashion items.\par
 These old approaches, though accurate, only have a small number of garment types, and thus is not practically useful on e-commerce websites. Our project leverages ModaNet dataset \cite{ModaNet} and uses Mask-RCNN to perform in parallel both recognition and segmentation of 10+ different types of fashion items. 
\subsection{Clothing Styles \& Texture Switching}
In order to switch styles of items in the image, the previous attempts mainly used GAN \cite{NIPS2014_5423} based approaches. For instance, the study \cite{DBLP:journals/corr/IsolaZZE16} proposed one approach for image-to-image translation using conditional GANs, which transform an input image to another one with a different representation. Other methods include using Cycle-Consistent Adversarial Networks \cite{DBLP:journals/corr/ZhuPIE17} to transfer the texture of fashion items from one style to another.\par
While images generated by those GAN based approaches \cite{DBLP:journals/corr/MaJSSTG17, DBLP:journals/corr/Al-HalahSG17, DBLP:journals/corr/ChenK17aa} are more photo-realistic, they have several disadvantages. For every two textures, a new model has to be trained, which is time consuming and notoriously inefficient. In addition, the GAN generated patches can lose folds or wrinkles of the original items. Plus, the whole image not only the selected fashion items will be modified. We propose to firstly use Mask R-CNN to find where the selected fashion item is, so that other parts of the image will remain intact. We then use method based on Neural Style Transfer \cite{DBLP:journals/corr/GatysEB15a} to perform texture and style transformation in order to maintain the information of wrinkles and folds.
\subsection{Virtual Try-on}
In the context of virtual try-on, earlier approaches relied on 3D information of body shapes. Guan et al. \cite{Guan:2012:DDA:2185520.2185531} proposed DRAPE to simulate 2D clothing designs on 3D bodies in different shapes and poses. Eisert et al. \cite{Hilsmann:2009:TRC:1560058.1560067} retextured the garment dynamically based on a motion model from real-time visualization in a virtual mirror environment. Pons-Moll et al. \cite{Pons-Moll:2017:CSC:3072959.3073711} utilized a multipart 3D model of clothed bodies for clothing capture and  retargeting. Compared with other pure 2D image based methods without relying 3D information, those 3D measurements are computationally inefficient and are unrealistic for normal users.\par
For 2D implementations, Zhu et al. \cite{DBLP:journals/corr/abs-1710-07346} proposed one approach which decomposes the generative process into two condition stages by 1) generating a plausible semantic segmentation map; 2) using a generative mode with a mapping to render the final image with precise regions and texture conditioned on this map. Jetchey and Bergmann \cite{inproceedings:Jetchev} proposed a conditional analogy GAN to swap fashion articles, without other descriptive person representation. However, their models did not take pose variant into consideration and required paired images of clothes and a wearer.\par
Our work applies 2D image-based approach with our own methods (see section 3).

\section{Methods}
The method we propose has two components, Mask R-CNN \cite{maskrcnn} and Neural Style Transfer. "Virtual Fitting Room" firstly uses Mask R-CNN to find the regions of different fashion items, and secondly uses Neural Style Transfer to change the style of the selected fashion items (Figure 1). Both of the two components will be explained in details in this section.
\begin{figure}
\begin{center}
\includegraphics[height=1.3in]{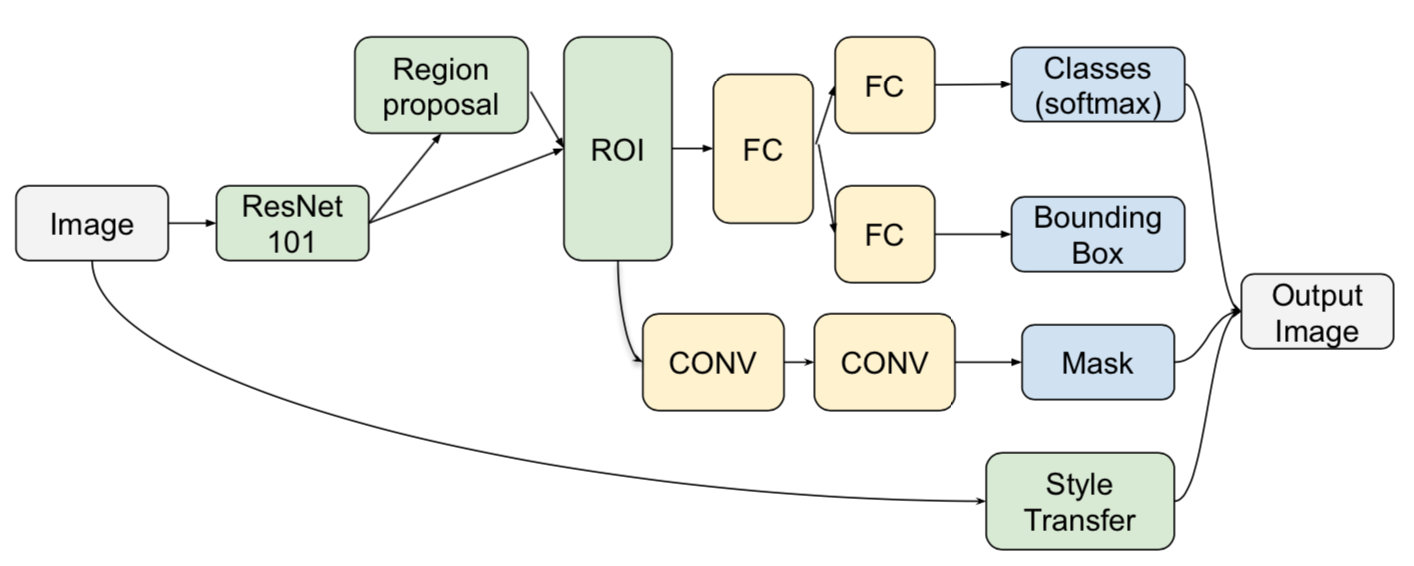}
\end{center}
   \caption{Virtual Fitting Room Architecture. FC represents fully connected layer; CONV represents convolutional layer.}
\label{fig:short}
\end{figure}

\subsection{Mask R-CNN}
Mask R-CNN uses a two-stage procedure. The first stage is the Regional Proposal Network which generates Region of Interest. In the second stage, the model, in parallel, predicts the class, box offset, and a binary mask. During training, we defined a multi-task loss on each sampled RoI as the sum of regional proposal network classification loss, regional proposal network bounding-box loss, mask classification loss, mask bounding-box loss and mask loss. These terms will be explained in details in Section 3.1.1 and 3.1.2. We used ResNet-101-FPN as the backbone of our network. The weights are pretrained with COCO or ImageNet. \par
As an operation for extracting a small feature map from each Region of Interest, RoIAlign is preferred here over RoIPool because RoIPool causes misalignments between the RoI and the extracted features. 
\subsubsection{Regional Proposal Network (RPN)}
Regional proposal networks propose Region of Interest. Each anchor box is given a binary class label. A positive label is given to two types of anchors: firstly the anchor with the highest Intersection-over-Union (IoU) overlap with a ground-truth box, or secondly an anchor that has an IoU higher than 0.7 with any ground-truth box. Loss function of an image is defined as:
\[L(\{p_{i}\}, \{t_{i}\}) = \frac{1}{N_{cls}}\sum_{i}L_{cls}(p_{i}, p_{i}^{*})\]
\[ + \lambda \frac{1}{N_{reg}}\sum_{i}p^{*}_{i}L_{reg}(t_{i}, t_{i}^{*}) \] 
\(N_{cls}\) is the number of images in a mini-batch; \(N_{reg}\) is the number of anchor locations in a mini-batch. i is the index of an anchor in a mini-batch. \(p_{i}\) is the predicted probability that anchor i is an object; \(p_{i}^{*}\) is ground truth, 1 if positive, and 0 otherwise. \(t_{i}\) is a vector of length 4 representing the predicted bounding box; \(t_{i}^{*}\) is the ground truth bounding box. \(\lambda\) is a balancing parameter between the two components. \(L_{cls}\) is the negative log loss function over two classes (object \(vs.\) not object); \(L_{reg}\) is the smooth \(L_{1}\) loss function. The \(p_{i}^{*}L_{reg}\) term means the bounding box regression loss is only activated for positive anchors, where \(p_{i}^{*}\) is 1.
We use a vector of length 4 to represent a bounding box:
\[t_{x} = (x-x_{a})/w_{a},\;\; t_{y} = (y-y_{a})/h_{a},\]
\[t_{w} = log(w/w_{a}),\;\; t_{h} = log(h/h_{a})\]
\[t_{x}^{*} = (x^{*}-x_{a})/w_{a},\;\; t_{y}^{*} = (y-y_{a})/h_{a},\]
\[t_{w}^{*} = log(w^{*}/w_{a}),\;\; t_{h}^{*} = log(h/h_{a})\]
\(x\), \(y\), \(w\), and \(h\) denote the predicted bounding box's center and its width and height. \(x^{*}\), \(y^{*}\), \(w^{*}\), and \(h^{*}\) denote the ground truth bounding box's center and its width and height. \(x_{a}\), \(y_{a}\), \(w_{a}\), and \(h_{a}\) denote the anchor box's center and its width and height.
\[L_{reg}(t, t^{*}) = \sum_{i\in{x,y,w,h}}smooth_{L_{1}}(t_{i} - t_{i}^{*})\]
in which
\[smooth_{L_{1}}(x) = \begin{cases}
                      0.5x^{2} & \text{if \abs{x} \textless 1} \\
                      \abs{x}-0.5 & \text{otherwise}
                      \end{cases}\]
\(L_{1}\) loss is less sensitive to outliers than the \(L_{2}\) loss, so \(L_{1}\) loss is more preferred here.

\subsubsection{Instance Segmentation}
\begin{figure}[h]
   \centering
   \includegraphics[height=1.2in]{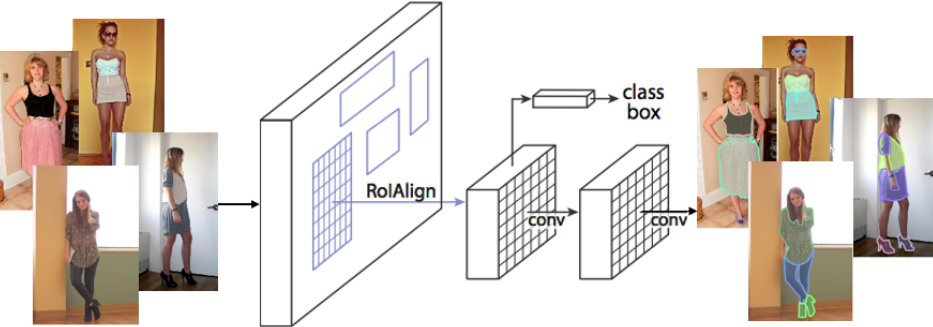}
   \caption{Instance Segmentation} 
   \label{fig:mask rcnn}
\end{figure}
When creating mask, we decouple class prediction and mask prediction (Figure 2) by creating bounding-box classification and mask in parallel. Mask classification loss and mask bounding box regression loss use same formulas as in Section 3.1.1. In this section, we only address mask loss. The dimension of the output of the mask branch is \(Km^{2}\), where \(K\) is the number of classes, and \(m*m\) is the resolution of the mask. After applying sigmoid function to the \(Km^{2}\) tensor, we define mask loss, \(L_{mask}\), as the average binary cross-entropy loss across the \(m*m\) pixels. For example, if \(k\) is the ground truth class, \(L_{mask}\) is the binary cross-entropy loss between the k-th predicted \(m*m\) mask and the ground truth mask. Predicted mask outputs of other classes do not contribute to \(L_{mask}\). When doing inference, we only use the k-th mask, where k is the predicted class by the classification branch. The k-th mask, which is of shape \(m*m\) will be resized to the RoI size, and be converted to 1 if bigger than 0.5, and 0 otherwise.

\subsection{Neural Style Transfer (NST)}
We used fashion images as content inputs and textures as style inputs. The produced images are fashion images content with artistic style of given texture. We implemented the style transfer technique from "Image Style Transfer Using Convolutional Neural Networks" \cite{nst}. SqueezeNet \cite{DBLP:journals/corr/IandolaMAHDK16}, a small model that has been trained on COCO or ImageNet, was chosen as the feature extractor for its small size and efficiency.\par
We performed gradient descent on the pixel values of our original image.
The loss function is a weighted sum of three terms: \textbf{content loss}, \textbf{style loss} and \textbf{total variation loss}. 

{\bf CONTENT LOSS} measures how much the feature map of the generated image differs from the feature map of the source image. We only care about the content representation of one layer of the network (say, layer \(\ell\)), that has feature maps \(A^{\ell} \in \mathbb{R}^{1\times C_{\ell} \times H_{\ell} \times W_{\ell}}\). \(C_{\ell}\) is the number of filters/channels in layer \(\ell\), \(H_{\ell}\) and \(W_{\ell}\) are the height and width. Let \(F^{\ell} \in \mathbb{R}^{C_{\ell} \times M_{\ell} }\) be the feature map for the current image and \(P^{\ell} \in \mathbb{R}^{C_{\ell} \times M_{\ell} }\) be the feature map for the content source image where \(M_{\ell} = H_{\ell} \times W_{\ell}\) is the number of elements in each feature map. Each row of \(F_{\ell}\) or \(P_{\ell}\)represents the vectorized activations of a particular filter, convolved over all positions of the image. \(i\) is in the range of \([0, C_{\ell}-1]\). \(j\) is in the range of \([0, M_{\ell}-1]\).  \(F_{ij}^{\ell}\) represents an element of \(F^{\ell}\) at \((i,j)\). \(P_{ij}^{\ell}\) represents an element of \(P^{\ell}\) at \((i,j)\). Finally, let \(w_{c}\) be the weight of the content loss term in the loss function. The content loss is given by:
\[L_{c} = w_{c} \times \sum_{i,j} (F_{ij}^{\ell} - P_{ij}^{\ell})^2,\]

{\bf STYLE LOSS} is defined as follows for a given layer \(\ell\):
We computed the Gram matrix G which represents the correlations between the responses of each filter, where F is as above. The Gram matrix is an approximation to the covariance matrix -- matching the (approximate) covariance is to make the activation statistics of our generated image match the activation statistics of our style image. Given a feature map \(F^{\ell}\) of shape \((C^{\ell}, M^{\ell})\), \(i\) and \(j\) are in the range of \([0, C_{\ell}-1]\). \(k\) is in the range of \([0, M_{\ell}-1]\).  \(F_{ik}^{\ell}\) represents an element of \(F^{\ell}\) at \((i,k)\). \(F_{jk}^{\ell}\) represents an element of \(F^{\ell}\) at \((j,k)\). The Gram matrix has shape \((C^{\ell}, C^{\ell})\) and its elements are given by:
\[G_{ij}^{\ell} = \sum_{k} F_{ik}^{\ell}  F_{jk}^{\ell}\]
 
Assuming \(G^{\ell}\) is the Gram matrix from the feature map of the current image, \(A^{\ell}\) is the Gram Matrix from the feature map of the source style image, and \(w_{\ell}\)  a scalar weight term, then the style loss for the layer \(\ell\) is the weighted Euclidean distance between the two Gram matric
\[L_{S}^{\ell} = w_{\ell} \sum_{i,j} (G_{ij}^{\ell} - A_{ij}^{\ell})^2\]

We computed the style loss at a set of layers \(\mathcal{L}\), then the total style loss is the sum of style losses at each layer:

\[L_{S} = \sum_{\ell\in\mathcal{L}}L_{S}^{\ell} \]

{\bf TOTAL-VARIATION LOSS} is to penalizes "total variation" in the pixel values. It turns out that it's helpful to encourage smoothness in the image.

We computed the "total variation" as the sum of the squares of differences in the pixel values for all pairs of pixels that are next to each other. \(x\) is currently generated image. \(H\) and \(W\) are height and width of the image. \(x_{i,j,c}\) is the pixel value of currently generated image at pixel \((i,j)\) of channel \(c\). The total-variation regularization for each of the input channels, and weight the total summed loss by the total variation weight \(w_{t}\) :
\[L_{tv} = w_{t} \times (\sum_{c=1}^3 \sum_{i=1}^{H-1}\sum_{j=1}^{W}(x_{i+1,j,c} - x_{i,j,c})^2 \]
\[+ \sum_{c=1}^3 \sum_{i=1}^{H}\sum_{j=1}^{W-1}(x_{i,j+1,c} - x_{i,j,c})^2)\]

\section{Dataset and Preprocessing}
\begin{table*}[!htbp]
\begin{center}
\begin{tabular}{|l||c||c|c|c|c|c|}
\hline
\textbf{image\_id} & \textbf{id} & \textbf{bbox} & \textbf{category\_id} & \textbf{iscrowded} & \textbf{segmentation} \\
\hline\hline
\textbf{736791}&\textbf{0}&[160,247,97,18]&2&0&[[161,248,170,248,173,249,189,251,199,...\\
& \textbf{1}&[287,537,30,42] & 4 & 0&[[198,543,197,551,198,556,199,565,200,...\\
& \textbf{2}&[287,537,26,42] & 4 & 0&[[289,554,287,564,287,573,287,577,292,...\\
\hline
\end{tabular}
\end{center}
\caption{Structure of the ModaNet annotations, including image id, annotation id, bounding boxes, item category id, iscrowded and segmentation.}
\end{table*}
\begin{figure}[!htbp]
\begin{center}
   \includegraphics[height=2.0in]{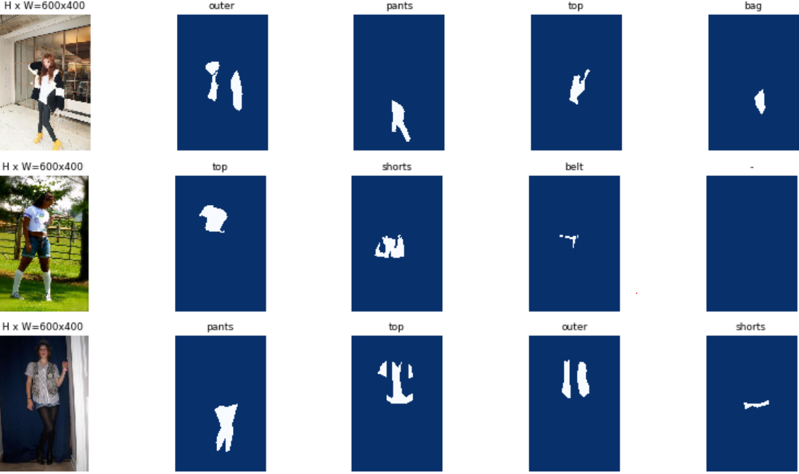}
  \caption{The first column displays raw image data from PaperDoll dataset, the right columns display the annotations retrieved from ModaNet}
\end{center}
\label{fig:long}
\label{fig:onecol}
\end{figure}
The data we used are composed of two parts, images and annotations. The raw image data comes from PaperDoll \cite{paperdollgithub} dataset, which is a collection of images of street-fashion models. Annotations are provided by eBay's ModaNet \cite{ModaNet}, which labels the PaperDoll dataset. The labels include bounding boxes and segmentations of 13 categories such as bag, belt, boots, top, shorts, scarf, tie etc defined by modanet.\par
The PaperDoll dataset images are stored in LMDB as encoded binaries. The key is photo id and the value is raw PNG binary. \par
ModaNet is the annotation of RGB images in PaperDoll dataset. ModaNet provides multiple polygon annotations for each image (Figure 3). Each polygon is associated with a label from 13 meta fashion categories. Table 1 shows the structure of the ModaNet annotation file. The second column 'id' is the annotation id. In the example of Table 1, there are 3 annotations. The first annotation in the table has \(category\_id\) 2. It means the annotation is a belt, according to ModaNet annotation guidelines. The label is in COCO \cite{COCO} style. The COCO bounding box format is [top left x position, top left y position, width, height]. "iscrowd" indicates whether the annotated item is crowded or not. "segmentation" provides a list of coordinates at the edge of the annotated fashion item.\par
We used 20,000 images for training, 2,000 images for validation and another 1,000 images for testing. The resolution of the image is around \(600*400\), except a few outliers. Each image is then resized to \(256*256\). We padded the image with zeros (black) on top and bottom, or left and right, if the image is not of square shape.\par
We did \textit{channel-level normalization} based on mean and variance calculated with ImageNet or Coco data set. We also did \textit{data augmentation} by flipping the image horizontally.
\section{Experiments}
\begin{table*}[!htbp]
\begin{center}
\begin{adjustbox}{max width=\textwidth}
\begin{tabular}{|c||c||c|c||c|c||c|c||c|c||c|c|}
\hline
\multicolumn{1}{c}{\textbf{Model}} & \multicolumn{1}{c}{\textbf{Preloaded}} & \multicolumn{2}{c}{\textbf{Epoch 1-50}} & \multicolumn{2}{c}{\textbf{Epoch 51-100}} & \multicolumn{2}{c}{\textbf{Epoch 101-150}} & \multicolumn{2}{c}{\textbf{Epoch 151-200}} & \multicolumn{1}{c}{\textbf{mAP(\%)}}\\ \hline
\multicolumn{1}{c}{} & \multicolumn{1}{c}{\textbf{}} & \multicolumn{1}{c}{\textbf{Layers1}} & \multicolumn{1}{c}{\textbf{LR1}} & \multicolumn{1}{c}{\textbf{Layers2}} & \multicolumn{1}{c}{\textbf{LR2}} & \multicolumn{1}{c}{\textbf{Layers3}} & \multicolumn{1}{c}{\textbf{LR3}} & \multicolumn{1}{c}{\textbf{Layers4}} & \multicolumn{1}{c}{\textbf{LR4}} & \multicolumn{1}{c}{\textbf{}}\\
M1 & ImageNet & All & 5e-4 & All & 5e-4 & All & 5e-5 & All & 5e-5 &41.62 \\
M2 & COCO & All & 5e-4 & All & 5e-4 & All & 5e-5 & All & 5e-5 &56.75 \\
M3 & ImageNet & Heads & 5e-4 & Heads & 5e-4 & All & 5e-5 & All & 5e-5 &48.61 \\
M4 & COCO & Heads & 1e-3 & Heads & 1e-3 & All & 1e-4 & All & 1e-4 &60.28 \\
\textbf{M5} & COCO & Heads & 1e-3 & C4, C5, Heads & 1e-3 & C4, C5, Heads & 1e-3 & All & 1e-4 &\textbf{68.72} \\
M6 & COCO & Heads & 1e-3 & Heads & 1e-3 & C5, Heads & 5e-4 & C5, Heads & 2e-4 &58.61 \\
M7 & COCO & Heads & 1e-3 & Heads & 1e-3 & C5, Heads & 1e-4 & All & 1e-4 &64.78 \\
M8 & COCO & Heads & 1e-3 & Heads & 1e-3 & Heads & 1e-4 & Heads & 1e-4 &50.09 \\
\textbf{FCN-CRF} &-  &-  &-  &-  &-  &-  &-  &-  &-  &\textbf{66.70}\\
\textbf{PaperDoll} &-  &-  &-  &-  &-  &-  &-  &-  &-  &\textbf{33.34}\\ \hline
\end{tabular}
\end{adjustbox}
\end{center}
\caption{We trained 8 models, M1, M2 ... M8. The Layers columns describe the trained layers in each step of the training process; The LRs columns describe the different learning rate we used. In the last column, we compared the mAP among all our models and the baseline models. Heads: Mask R-CNN heads, Regional Proposal Network heads and Feature Pyramid Network heads, as discussed in Section 3; C4, C5: the \(4th\) and \(5th\) component in ResNet-101-FPN \cite{fpn}; All: all layers in our network; FCN-CRF \cite{baseline1}, PaperDoll Parsing \cite{baseline2}: baseline models.}
\end{table*}
As discussed in Section 3, our software is mainly composed of two parts, Mask R-CNN and Neural Style Transfer. Since these two components are independent, in this section, we will firstly discuss the results of Mask R-CNN and Neural Style Transfer separately. And at the end of this section, we will present the final results, which is a merge of the outputs from Mask R-CNN and Neural Style Transfer.
\subsection{Mask R-CNN Training Details}
When training Mask R-CNN, there are several parts where we can tune: 1) different initialization, either ImageNet or Coco, 2) different trainable layers, any combinations of Mask R-CNN heads, RPN heads, Feature Pyramid Network (FPN) heads, and layers in ResNet, 3) different training schedules, where for example we can train Mask R-CNN heads only for the first 100 epochs and then train all layers for another 100 epochs. The results of these different experiments are in Section 5.3.\par
Minibatch size is 32. The reason that we pick 32 is that it is the largest batch that we are able to fit into a K80 GPU. We decrease learning rate as we increase trainable layers in later training steps. For example, in model M3 (Section 5.3), we used 0.001 as learning rate when we firstly train Mask R-CNN head, RPN head, and FPN head for the first 100 epochs, and then use 0.0001, which is 10 times smaller, as we fine-tune all layers. The reason is that when we do fine-tuning, we do not want to update the parameters in ResNet too aggressively, because we expect these weights to be quite good since they are initialized with weights trained on COCO or ImageNet. A learning rate that is too big will likely distort these parameters.\par
\begin{figure*}[!htbp]
   \centering
   \includegraphics[height=2.2in]{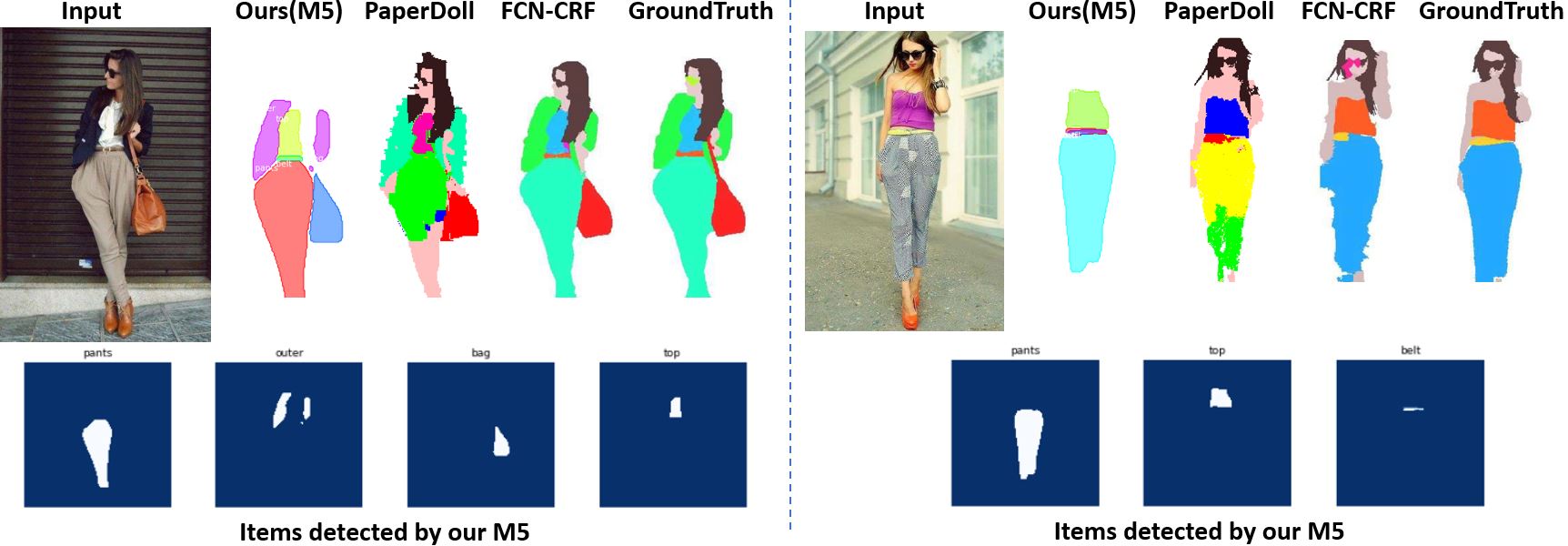}
   \caption{Comparison of segmentation between M5, the two baseline models (FCN-CRF \cite{baseline1} and PaperDoll Parsing \cite{baseline2}) and ground truth. We chose M5 among the 8 trained models because M5 has the highest mAP.}
   \label{fig:transfer results}
\end{figure*}
\subsection{Segmentation Quantitative Metric: mAP}
\[Precision = \frac{True Positive}{True Positive + False Positive}\]
\[Recall = \frac{True Positive}{True Positive + False Negative}\]
We can get a list of precisions and recalls for each category of fashion items, at different class score thresholds. We calculate the Average Precision for each cateogry, which is the area under the precision recall curve (r is recall):
\[AP = \int_{0}^{1} p(r) dr\]
We padded precisions with 1 at the beginning and 0 at the end; recalls with 0 at the beginning and 1 at the end, for ease of calculation. We approximated the integral with:
\[AP = \sum_{i=1}^{l-1} (recalls[i] - recalls[i-1]) * precisions[i]\]
where i is index of the list of ascending recalls from 0 to 1; l is length of the list of recalls.\par
After we have computed the AP of each category, mAP is calculated by taking the mean AP over all categories.
\subsection{MRCNN \& Baseline Quantitative Comparison}
\begin{figure}[!htbp]
\begin{center}
   \includegraphics[height=2.4in]{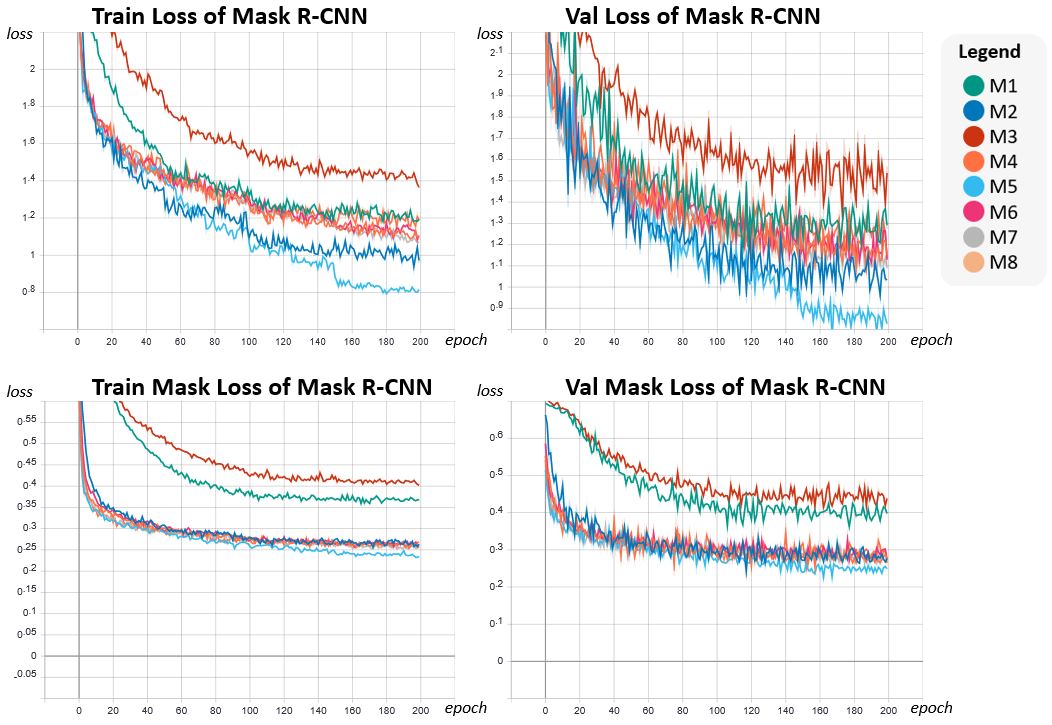}
  \caption{Validation total loss and validation mask loss keep decreasing. The gap between training loss and validation loss is also small. There is no sign of overfitting.}
\end{center}
\label{fig:long}
\label{fig:onecol}
\end{figure}\par
We trained 8 different models and compared the mAP of our models with that of FCN-CRF \cite{baseline1} and PaperDoll Parsing \cite{baseline2}. We recorded the results in Table 2 and made several observations. \par
Firstly, M5 and M7 achieved the two highest mAP. The reason is that M5 and M7 adopted a 3-steps training. M5 trained on Heads only for the first 50 epochs, then C4, C5 (see description in the table 2 caption) and Heads for the next 100 epochs and lastly all layers for 50 epochs; M7 has similar training schedules. M5 and M7 did the best in taking advantage of the pre-trained weights in our backbone network. At the beginning of the training, they both froze the pre-trained weights to avoid distortion of the backbone network's weights. It also made the training of the following part of the network more efficient because it can rely on the meaningful feature extraction by the pre-trained backbone network. During the training process, M5 and M7 both gradually increase the number of trainable layers, instead of suddenly starting to train all layers, such as M3 and M4. It helped reduce the risk of distortion of pre-trained weights.\par
Secondly, M1 and M2, that started training all layers from the very beginning, performed very badly. The reason is that training the backbone weights while the weights of the following network are random and meaningless, which happens at the beginning of the training process, will very likely distort the pre-trained weights.\par
Thirdly, it is important to train all layers. The reason that M6 and M8 have suboptimal performance is that they only trained parts of the model, without fine-tuning all layers. To suite our specific purpose, even the first several lalyers of the backbone network must be trained. \par
Fourthly, our models did not overfit the training data set. We used L2 weights regularization to mitigate the risk of overfitting. The validation loss of our models kept decreasing and reached a plateau (Figure 5). The validation loss never increased at the end of our training process. The gap between validation loss and training loss is small. The high mAP scores also prove that our models generalize well to unseen data. M5, which has the highest mAP, also has the lowest validation total loss and validation mask loss.
\subsection{MRCNN \& Baseline Qualitative Comparison}
We compared the segmentation results of M5 qualitatively with those of FCN-CRF\cite{baseline1} and PaperDoll Parsing\cite{baseline2} (Figure 4). The results of M5 is apparently better than those of FCN-CRF and PaperDoll Parsing. The edges of the segmentation of both PaperDoll Parsing and FCN-CRF are not smooth. The coarse outline of segmentation is fine if the goal is classification. However for virtual try on, we need a clear and smooth outline of each fashion item, so that the item can be cleanly substituted. M5 outperformed both baseline models, as expected, because M5 has a much higher mAP.
\subsection{Neural Style Transfer Results}
We transferred various textures onto portrait images and got decent results (Figure 6). The textures we tried include jeans, leather, cloud, composition and muse. The optimal hyper-parameters (Figure 6) of the Neural Style Transfer network are different for different textures. We tuned the hyper-parameters based on qualitative results.
\begin{figure}[!htbp]
\begin{center}
   \includegraphics[height=3in]{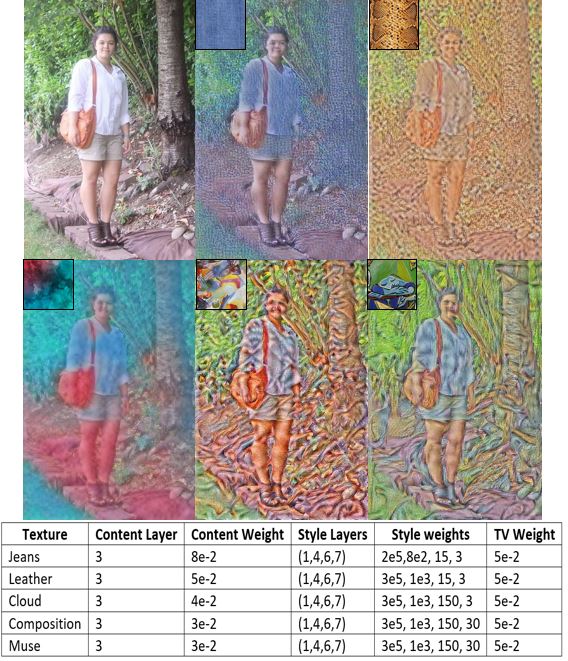}
  \caption{Image 1 in top left is the original image. We used jeans, leather, cloud, composition and muse as style input. The small boxes on the top left corner of each image is the style/texture we want to transfer to the original image. Image 2-6 are the results. The corresponding texture and with tuned hyper-parameters are also listed. 'TV' represents for total variation.}
\end{center}
\label{fig:long}
\label{fig:onecol}
\end{figure}

\subsection{Final Results Quantitative Metric}
We used mAP as a quantitative metric to evaluate Mask R-CNN results (Section 5.3). For final results, we use Average Score Decay Rate to evaluate.
\[ASDR = \frac{1}{n} \sum_{i=1}^{n}\frac{max(S^{B}_{i}-S^{A}_{i}, 0)}{S^{B}_{i}}\]
n is the total number of fashion items in test set. \(S_{i}^{B}\) is the confidence score that the \(ith\) fashion item exists in the original image before style transfer. \(S_{i}^{A}\) is the confidence score that the \(ith\) fashion item exists in the new image after style transfer. The logic behind ASDR is quite intuitive. For example, if a dress appears in the original image, a dress should still appear in the new image after the style transfer. \textbf{A smaller ASDR means a better performance.}
\subsection{Final Results \& Baseline Comparison}
We compared the results of our models (Figure 7). We chose two segmentation models with the highest mAP, M5 and M7. "M5+NST" uses M5 to find the selected fashion item and then uses neural style transfer (NST)  to change the style of the selected fashion item. "M7+NST" uses M7 instead. "M5+CopyPaste" uses M5 and then copy and paste the new texture directly onto the fashion item.\par
"M5+NST" apparently has the best results among our models. Compared to "M7+NST", "M5+NST" has more accurate segmentation. For instance, on row two column one, "M7+NST" failed to find the complete outer and overlooked a small piece of that white outer. The consequence is that the overlooked small piece of outer did not have a style change and remained to be white. Compared to "M5+CopyPaste", "M5+NST" is able to create new fashion items whose brightness and color fit better into the original image.\par
\begin{figure}[!htbp]   
\begin{center}
   \includegraphics[height=2.6in]{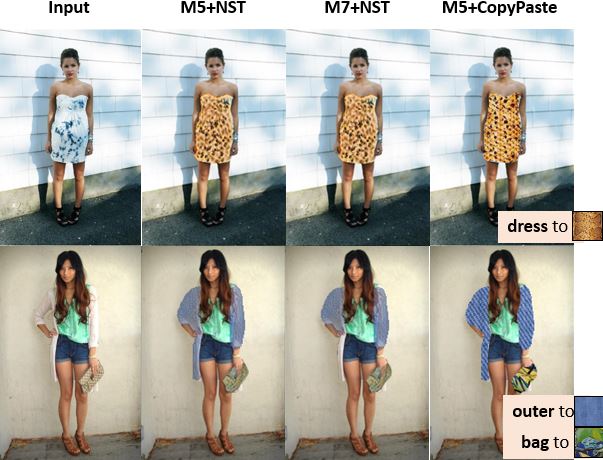}
  \caption{"M5+NST": use M5 to find the selected fashion item and then apply NST to change the style of the fashion item; "M7+NST": uses M7 instead; "M5+CopyPaste" uses M5 to find the selected fashion item and then copy and past the new texture onto the selected fashion item. For the first image, we changed the texture of the dress to snake leather; For the second image, we switched the style of the outer to jean and the style of the bag to muse.}
  \end{center}
\label{fig:long}
\label{fig:onecol}
\end{figure}
We compared the results of our method with a number of baselines (Figure 8). From the results, the PRGAN \cite{DBLP:journals/corr/MaJSSTG17} and Encoder-Decoder were able to locate the region of the top correctly and generate texture to the target position. However, the results generated by those two methods are blurry and coarse. The details of the target texture are missing. The CAGAN \cite{inproceedings:Jetchev} succeeded in rendering the details of the texture but also cause undesirable artifacts. For instance, some edges of the clothing item are blur; the bottom part of the tops are not well rendered. All those three methods also change the color of the trousers which should not be changed. In contrast, our method effectively rendered the target texture on to the top of the input image without changing other areas of the image.
\begin{figure*}[!htbp]
\begin{center}
   \includegraphics[height=2.2in]{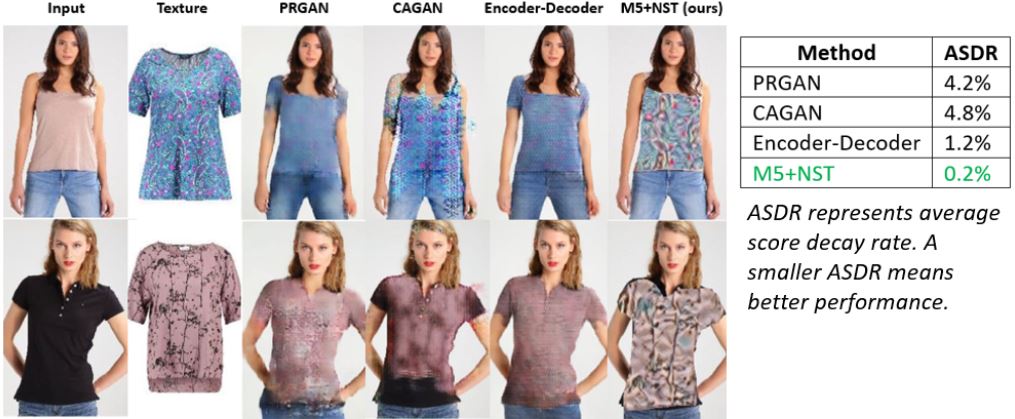}
  \caption{Qualitative comparison between our methods and others. The first column displays input images. The second column displays the target textures to apply. We compare our final result by M5 and Neural Style Transfer (most right column) with PRGAN \cite{DBLP:journals/corr/MaJSSTG17}, CAGAN \cite{inproceedings:Jetchev} and Encoder-Decoder.}
\end{center}
\label{fig:long}
\label{fig:onecol}
\end{figure*}
\begin{figure}[!htbp]
\begin{center}
   \includegraphics[height=2.8in]{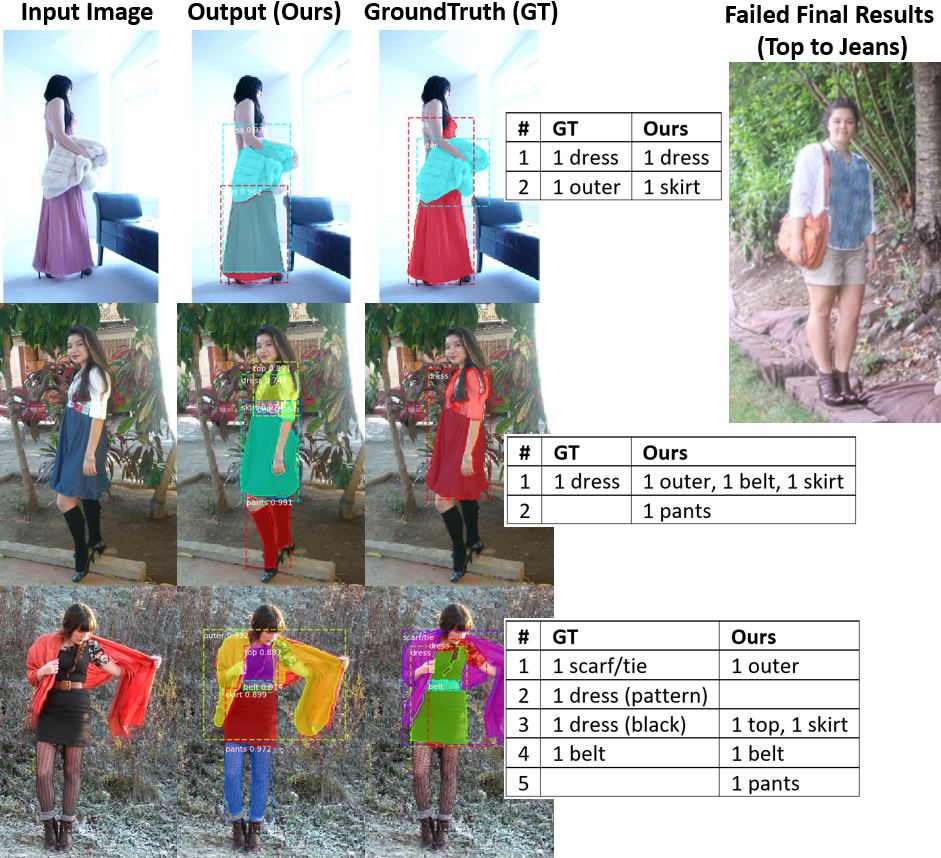}
  \caption{Examples of failed results. The first column displays input images. The second column displays our results of M5. The third column displays ground truth results(annotations retrieved from ModaNet). The top right image is one of the failed results after texture transfer.}
\end{center}
\label{fig:long}
\label{fig:onecol}
\end{figure}
\subsection{Failed Results Analysis}
Samples of failed final results are presented in Figure 9. The first column displays input image. The second column displays results of our M5. The third column displays ground truth results(annotations retrieved from ModaNet). In some failed cases, our model is fooled by a belt or color boundary. When a piece of clothes is visually divided into two parts by a belt, it can be detected as two clothes. Some footwear and boots are detected as pants because we removed the labels for boots and footwear during training for the purpose of simplicity at the current stage. \par
For the first example, the outer which is under the girl's arm visually divided the dress into two parts, so the dress is detected as skirt. For the second example, sharp boundary of two colors on the dress makes the dress look like a outer and skirt. Stockings were detected as pants because we removed the labels for boots and footwear during training. For the third example, the scarf looks like an outer. The dress is detected as top and skirt because it's visually divided into two parts by the belt. The silk stockings were detected as pants due to the same reason as second example.\par
For the failed final result case displayed top right in Figure 9, the model is fooled by the bag belt and thought the right part of the outer as top. 
\section{Conclusion and future work}
Virtual fitting room is a challenging task yet useful feature for e-commerce platforms and fashion designers. There are three key issues that make the problem difficult. Firstly there can be a big number of fashion items that are of different categories in one single fashion model image. Secondly it is hard to find a perfect segmentation for each fashion item. Thirdly it is even more difficult to change the fashion item into any arbitrary style.\par
We tried to solve the problem of virtual try on in two steps and saw good results. We used Mask R-CNN to find the regions of different fashion items, and Neural Style Transfer to change the style of the selected fashion items. To improve the performance of the model, We employed deep learning techniques such as transfer learning, fine tuning and hyper-parameter tuning. M5 has the best qualitative and quantitative result (68.72 mAP) because it has the most efficient training schedule to best take advantage of the initialization weights in backbone network. For future work, we would like to further increase the number of types of detectable fashion items and also transferable textures. This can be achieved by labeling more data.
\clearpage

\section{Contributions \& Acknowledgements}
Jie set up the Google Cloud environment, created the Git repository, did investigations on related works, processed the data and trained some models. Junwen focused on network modeling, explored the original model implementation and performed baseline comparison. Zhiling did the literature review and focused on dataset processing, performed transfer learning and fine tuning. All team members worked together to develop model's architecture, tune hyper-parameters, train models, analyze results and write final report.
\section{Starter Code}
Our Mask R-CNN implementation is based on \href{https://github.com/matterport/Mask\_RCNN}{https://github.com/matterport/Mask\_RCNN}. We created our own classes to train our fashion dataset. Overrode and rewrote the DataSet class and Configuration class. We created Notebooks specific for our project to perform test and visualize our results. We also modified the existing implementation in "mask\_rcnn.mrcnn" package to meet requirements specific for our project. Those modifications including selecting different layers for model training, some hyperparameters logging, and helper functions for visualization.\par
Our Neural Style Transfer implementation is based on "Style Transfer" part in assignment 3 of CS231n. We referred the content in Jupyter Note book and created our own script which accepts two input images and output texture images per 10 iterations so that we can compare the generated results per 10 iterations and perform hyperparameter tuning and select the best results for certain texture.\par
For more implementation details and training instructions, please refer our repository. Our repository can be found here: \href{https://github.com/jiechen2358/VirtualFittingRoom}{https://github.com/jiechen2358/VirtualFittingRoom}.

\section{Acknowledgement}
We would like to thank Pratyaksh Sharma for his helpful feedback and comments for our project. We would also like to thank the CS231n teaching team for their prompt answers to our questions on the Piazza forum. Finally, we would like to thank Google Cloud for providing us with credit.
{\small
\bibliographystyle{ieee}
\bibliography{main}
}

\end{document}